\documentclass[letterpaper, 10 pt, conference]{ieeeconf}

\IEEEoverridecommandlockouts                              

\usepackage{graphicx}
\usepackage{tabu}
\usepackage{amsmath,amssymb}
\usepackage{algorithm}
\usepackage[noend]{algpseudocode}
\usepackage{longtable}
\usepackage{mathtools}
\usepackage{breqn}
\usepackage{caption}
\usepackage{amsmath}
\DeclareMathOperator*{\argmax}{arg\,max}

\title{\LARGE \bf
Combining Deep Reinforcement Learning And Local Control \\
For The Acrobot Swing-up And Balance Task
}
\author{Sean Gillen, Marco Molnar, and Katie Byl
\thanks{*This work was funded in part by NSF NRI award 1526424.}
\thanks{Sean Gillen and Katie Byl are with the Electrical and Computer Engineering Department at the University of California, Santa Barbara CA 93106
{\tt\small sgillen@ucsb.edu},
{\tt\small katiebyl@ucsb.edu}.
Marco Molnar is with TU Berlin
{\tt\small marco.molnar@posteo.de}.}
\thanks{**Source code for this paper can be found here: https://github.com/sgillen/ssac}%
        }

\begin{document}

\DeclarePairedDelimiterX{\infdivx}[2]{(}{)}{%
  #1\;\delimsize\|\;#2%
}
\newcommand{\infdiv}{D\infdivx}
\DeclarePairedDelimiter{\norm}{\lVert}{\rVert}

\maketitle

\begin{abstract}
In this work we present a novel extension of soft actor critic, a state of the art deep reinforcement algorithm. Our method allows us to combine traditional controllers with learned neural network policies. This combination allows us to leverage both our own domain knowledge and some of the advantages of model free reinforcement learning. We demonstrate our algorithm by combining a hand designed linear quadratic regulator with a learned controller for the acrobot problem. We show that our technique outperforms other state of the art reinforcement learning algorithms in this setting.
\end{abstract}

\section{INTRODUCTION}

Advances in machine learning have allowed researchers to leverage the massive amount of compute available today in order to better control robotic systems. The result is that modern model-free reinforcement learning has been used to solve very difficult problems. Recent examples include controlling a 47 DOF humanoid to navigate a variety of obstacles \cite{heess_emergence_2017}, dexterously manipulating objects with a 24 DOF robotic hand \cite{openai_learning_2018}, and allowing a physical quadruped robot to run \cite{hwangbo_learning_2019}, and recover from falls \cite{lee_robust_2019}.

Despite this, these algorithms can struggle on certain low dimensional problems from the nonlinear control literature. Namely  the acrobot \cite{spong_swing_1994} and the cart pole pendulum. These are both under-actuated mechanical systems that have unstable fixed points in their unforced dynamics (see section \ref{section:Acrobot}). Typically, the goal is to bring the system to this fixed point and keep it there. In this paper we focus on the acrobot as we found less examples of model free reinforcement learning performing well on this task.

\begin{figure}[ht]
\centering
  \includegraphics[scale=.25]{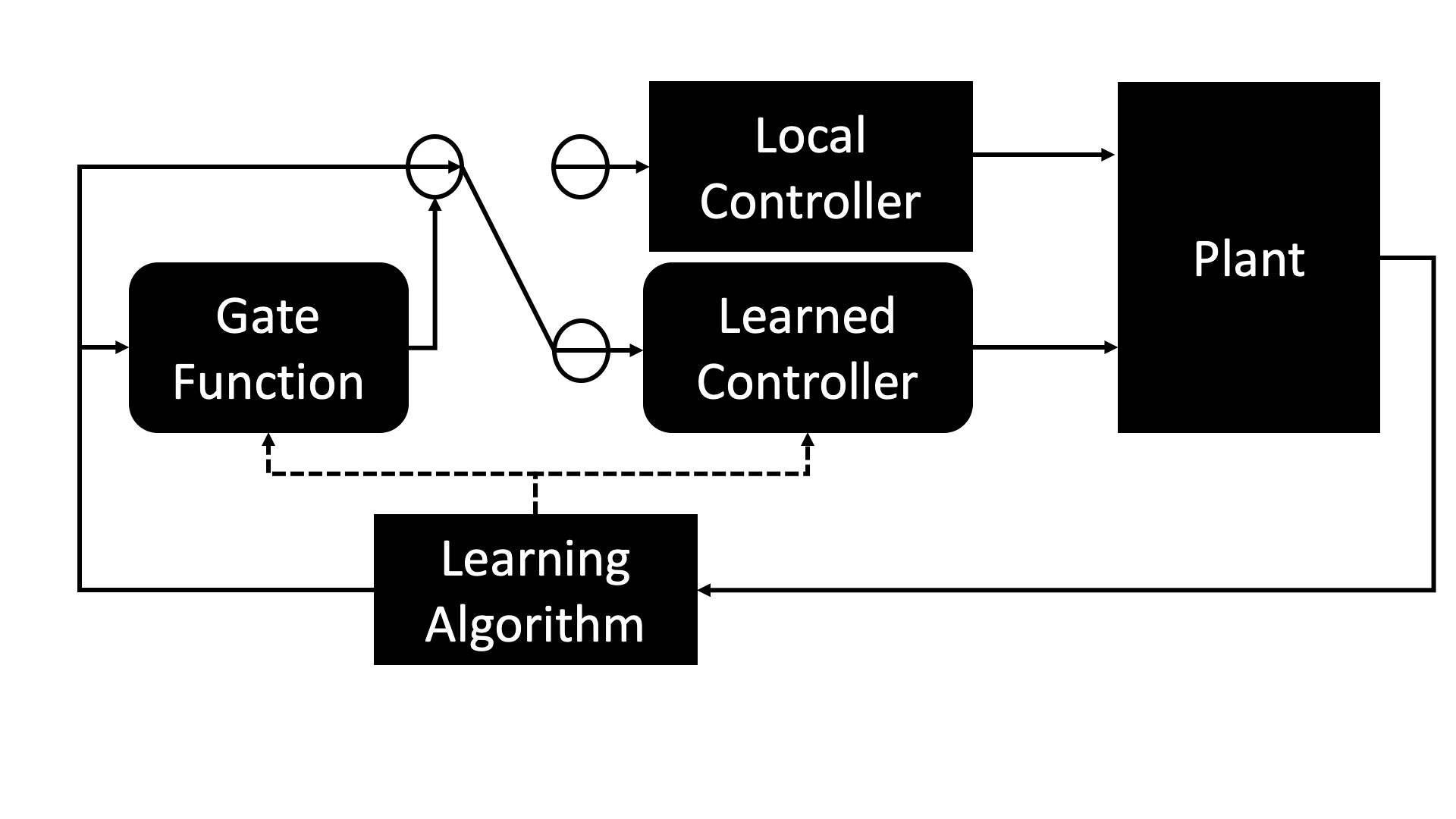}
  \caption{System diagram for the new technique proposed in this paper. Rounded boxes represent learned neural networks, squared boxes represent static, hand crafted functions. The local controller is a hand designed LQR, the swing-up controller is obtained via reinforcement learning, and the gating function is trained as a neural network classifier}
  \label{fig:hyst}
\end{figure}

It is not uncommon to see some variation of these systems tackled in various reinforcement benchmarks, but we have found these problems have usually been artificially modified to make them easier. For example the very popular OpenAI Gym benchmarks \cite{1606.01540} includes an acrobot task. But the objective is only to get the system in the rough area of the unstable fixed point, and the dynamics are integrated with a fixed time-step of .2 seconds, which makes the problem much easier and unrepresentative of a physical system. We have found that almost universally, modern model free reinforcement learning algorithms fail to solve a more realistic version of the task. Notably, the Deep Mind control suite \cite{deepmindcontrolsuite2018} includes the full acrobot problem, and all but one algorithm that they tested (the exception being \cite{barth-maron_distributed_2018}) learned nothing, the average return after training was the same as before training. 

Despite this, there are many traditional model based solutions \cite{spong_swing_1994}, \cite{spong_energy_1996}, that can solve this problem well. In this work we do not seek to improve upon the model based solutions to this problem, but to extend to the class of problems that model free reinforcement learning methods can be used to solve. We believe the methods used here to solve the acrobot can be extended to other problems, such as making robust walking policies.

 One of the primary reasons why this problem is difficult for RL is that the region of state space that can be brought to the unstable fixed point is very small, even with generous torque limits. An untrained RL agent explores by taking random actions in the environment. Reaching the region of attraction is rare, we found that for our system, random actions will reach the basin of attraction for a well designed LQR in about 1\% of trials. However an RL agent doesn't have access to a well designed LQR at the start of training, in addition to reaching the region where stabilization is possible, the agent must  also stabilize the acrobot for the agent to receive a strong reward signal. This results in successful trials in this environment being extremely rare, and therefore training is in-feasibly slow and sample inefficient.

Our solution to add a predesigned balancing controller into the system, this is comparatively easy to design, and can be done with a linear controller. Our contribution is a novel way to combine this balancing controller with an algorithm that is learning the swing-up behavior. We simultaneously learn the swing-up controller, and a function that switches between the two controllers.  

\subsection{Related Work}

Work done by Randolov et. al. \cite{randlov_combining_2000} is closely related to our own. In that work they construct a local controller, an LQR, and combine it with a learned controller to swing-up and balance an inverted double pendulum (similar to the acrobot we study but with actuators at both joints). The primary differences between our work and theirs is that they hard code the transition between their two controllers. In contrast we learn our transition function online and in parallel with our swing-up controller.


Work done by Yoshimoto et. al. \cite{yoshimoto_acrobot_2005}, like ours, learns the transition function between controllers in order to swing-up and balance an acrobot. However, unlike our work they limit the controllers they switch between to pre-computed linear functions. In contrast our work simultaneously learns a nonlinear swing-up controller and the transition between a learned and pre-computed balance controller.


Wiklednt et. al  \cite{wiklendt_small_2009} too swing-up and balance an acrobot using a combined neural network and LQR. However they only learn to swing-up from a single initial condition, whereas our method learns to solve the task from any initial position.


Doya \cite{doya_multiple_2002} also learns many controllers using reinforcement learning, and adaptively switches between them. However unlike our work, the switching function is not learned using reinforcement learning, but is instead selected according to which of the controllers currently makes best prediction of the state at the current point in state space. We believe our model free updates will avoid the model bias that can be associated with such approaches. Furthermore our work allows for combining learned controllers with hand designed controllers, such as the LQR.  

\section{Background}
\subsection{Nomenclature}

We formulate our problem as a Markov decision process, $M = (S,A,R) $. At each time step $t$, Our agent receives the current state $s_{t} \in S$ and chooses an action $a_{t} \in A$. It then receives a reward according to the reward function $r_{t} = R(s_{t}, a_{t}, s_{t+1})$. The goal is to find a policy  $\pi: S \rightarrow p(A = a | s)$ that satisfies:

 \begin{dmath} \pi^{*} = \argmax_{\pi} \mathbb{E}\left[ \sum_{t=0}^{\infty}\gamma^{t}R(s_{t}, a_{t}, s_{t+1}) \right]  \end{dmath}

\subsection{Acrobot}
\label{section:Acrobot}

\begin{figure}[ht]
\centering
  \includegraphics[scale=.25]{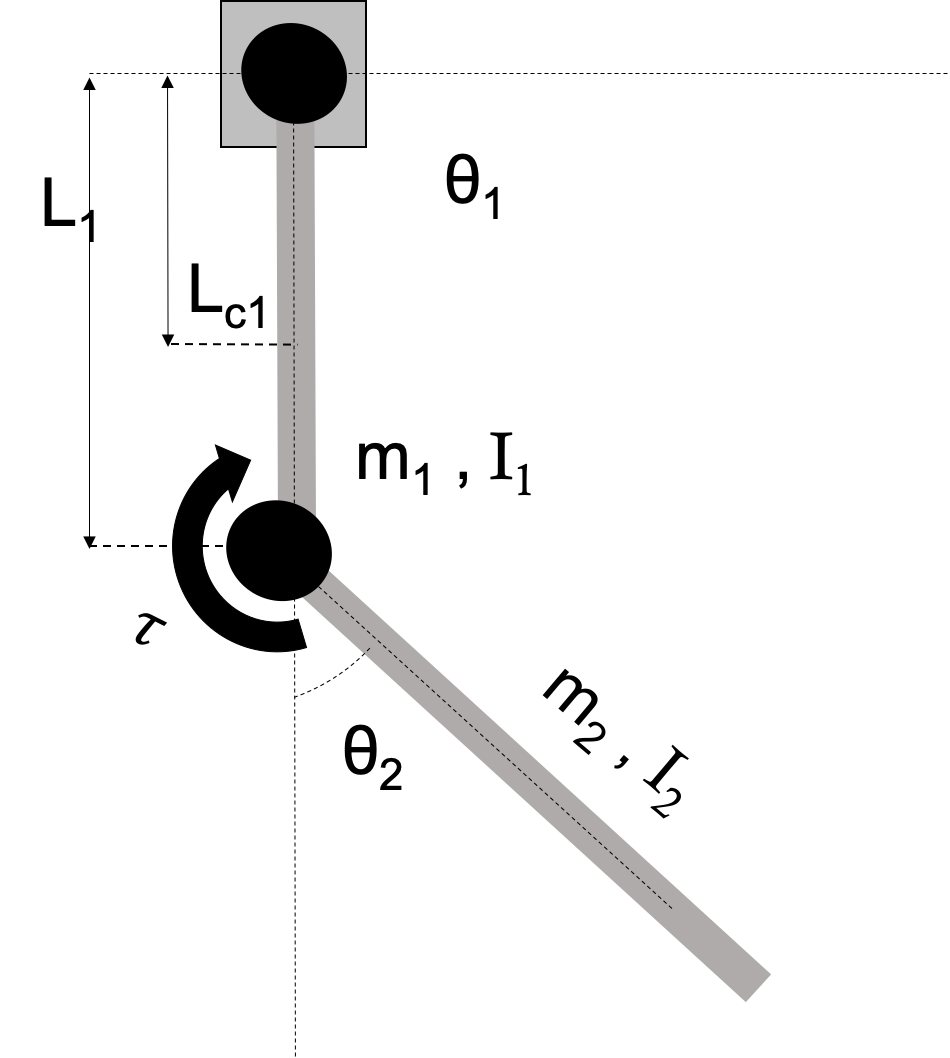}
  \caption{Diagram for the acrobot system}
  \label{fig:acrobot}
\end{figure}

The acrobot is described in Figure \ref{fig:acrobot}. It is a double inverted pendulum with a motor only at the elbow. We use the parameters from Spong \cite{spong_swing_1994}:

\begin{center}
\captionof{table}{Mass and inertial parameters used in simulation}
\begin{tabular}{ | c | c | c | }
\hline
Parameter & Value & Units\\
 \hline
 $m_{1}, m_{2}$ & 1 & Kg \\ 
 \hline
 $l_{1}, l_{2}$ & 1 & m  \\ 
 \hline
 $l_{c1}, l_{c2}$  & .5 & m  \\ 
 \hline
 $I_{1}$ & .2 & Kg*m$^{2}$ \\
 \hline 
 $I_{2}$ & 1.0 & Kg*m$^{2}$  \\ 
 \hline
\end{tabular}
\end{center}

The state of this system is $s_{t} = [\theta_{1}, \theta_{2}, \dot \theta_{1}, \dot \theta_{2}]$. The action $a_{t} = \tau$, is the torque at the elbow joint. The goal we wish to achieve is to take this system from any random initial state, to the upright state $gs = [\pi/2, 0, 0, 0]$, which we will refer to as the goal state. To achieve this goal, we seek the maximum of the following reward function:

\begin{dmath} r_{t} = l_{1}\sin(\theta_{1}) + l_{2} \sin(\theta_{1} + \theta_{2})
\end{dmath}
This was motivated by the popular Acrobot-v1 environment \cite{baselines}, We found empirically that for our algorithm this reward signal led to the same solutions as the more typical $\norm{s_{t} - gs}$. However we found that some of the other algorithms we compared to perform better with the sinusoidal reward function. 

We implement the system in python (all source code is provided, see footnote on page one), the dynamics are implemented using Euler integration with a time-step of .01 seconds, and the control is updated every .2 seconds. We experimented with smaller timesteps and higher order integrators, generally we found these made the balancing task easier, but made the wall clock time for the learning much slower.


\subsection{Soft Actor Critic}

Soft actor critic (SAC) is an off policy deep reinforcement learning algorithm shown to do well on control tasks with continuous actions spaces \cite{haarnoja_soft_2018}. To aid in exploration, rather than directly optimize the discounted sum of future rewards, SAC attempt to find a policy that optimizes a surrogate objective:

\begin{dmath}J^{\mathrm{soft}} = \mathbb{E}\left[ \sum_{t=0}^{\infty}\gamma^{t}\bigg(R_{t} + \alpha H(\pi(\cdot|s_{t}))\bigg) \right]  \label{eq:thresh} \end{dmath}
 
 Where $H$ is the entropy of the policy.
 



SAC introduces several neural networks for the training. We define a soft value function $V_{\phi}(s_{t})$, a neural network defined by weights $\phi$, which approximate $J^{soft}$ given the current state. Next we define two soft Q functions $Q_{\rho_{1}}(s_{t}, a_{t})$ and $Q_{\rho_{2}}(s_{t}, a_{t})$ which approximate $J^{soft}$ given both the current state and the current action. Using two Q networks is a trick that aids the training by avoiding overestimating the Q function. We must also define a target soft value function $V_{\overline \phi}(s_{t})$, which follows the value function via polyak averaging:

\begin{dmath}
V_{\overline \phi^{+}}(s_{t}) = c_{py}V_{\overline \phi}(s_{t}) +  (1 - c_{py})V_{\phi}
\end{dmath}

With $c_{py}$ a fixed hyper parameter. We also define $\Pi_{\theta}$, a neural network that outputs $\mu_{\theta}(s_{t})$ and $\log(\sigma_{\theta}(s_{t}))$ which define the probability distribution of our policy $\pi_{\theta}$. The action is given by:

\begin{dmath}
a_{t} = \tanh(\mu_{\theta}(s_{t}) + \sigma_{\theta}(s_{t}) \epsilon_{t})
\label{act}
\end{dmath}

where $\epsilon_{t}$ is drawn from $N(0,1)$. 
 
SAC also make use of a replay buffer $D$ which stores the tuple $(s_{t}, a_{t}, r_{t})$ after policy rollouts. When it is time to update we sample randomly from this buffer, and use those samples to compute our losses and update our weights.

With this we can define the losses for each of these networks (originated from \cite{haarnoja_soft_2018})
 
The loss for our two Q functions is:
 
\begin{dmath} L^{Q} = \mathbb{E}_{s_{t}, a_{t} \sim D}\left[\frac{1}{2}\left( Q_{\rho}(s_{t},a_{t}) - \hat{Q}(s_{t}, a_{t}) \right) ^{2}  \right] \end{dmath}
where
\begin{dmath} \hat{Q}(s_{t}, a_{t}) = r(s_{t}, a_{t}) + \gamma\mathbb{E}_{s_{t+1}}\left[V_{\overline{\phi}}(s_{t+1} ) \right]  \end{dmath}

Our policy seeks to minimize:

\begin{dmath} 
L^{\pi} = \mathbb{E}_{s_{t} \sim D, \epsilon_{t} \sim N(0,1)}\left[ \log \pi_{\theta}(f_{\theta}(\epsilon_{t}, s_{t}) | s_{t}) - Q_{\rho_{1}}(s_{t}, f_{\theta}(\epsilon_{t}, s_{t}) \right]
\end{dmath}

And our value function:

\begin{dmath}
L^{\text{V}} = \mathop{\mathbb{E}}_{s_{t} \sim D} \left[ \frac{1}{2} \left( V_{\phi}(s_{t}) - \hat  V_{\phi}(s_{t}) \right)^{2}  \right] 
\end{dmath} 

Where

\begin{dmath}
\hat V_{\phi} = \mathbb{E}_{a_{t} \sim \pi_{\theta}} \left[Q^{min}(s_{t}, a_{t}) - log \pi_{\theta}(a_{t} | s_{t}) \right]
\end{dmath}

And $Q^{min} = \min(Q_{\rho_{1}}(s_{t}, a_{t}), Q_{\rho_{2}}(s_{t}, a_{t}))$

 SAC starts by doing policy roll outs, recording the state, action, reward, and the active controller at each time step. It stores these experiences in the replay buffer. After enough trials have been run, we run our update step. We sample from the replay buffer, and use these sampled states to compute the losses above. We then run one step of Adam \cite{kingma_adam:_2014} to update our network weights. We repeat this update $n_{u}$ times with different samples. Finally we  copy our weights to our target network and repeat until convergence (or some other stopping metric).

\section{Switched Soft Actor Critic}
\label{section:SSAC}

Our primary contribution is to extend SAC in two key ways, we call the modified algorithm switched soft actor critic (SSAC). The first modification is a change to the structure of the learned controller in order to inject our domain knowledge into the learning. Our controller consists of three distinct components. The gate function, the balancing controller, and the swing-up controller. The gate, $G_{\gamma}: S \rightarrow [0,1]$, is a neural network parameterized by weights $\gamma$ which takes the observations at each time step and outputs a number $g_{t}$ representing which controller it thinks should be active. $g_{t} \approx 1$ implies high confidence that the balancing controller should be active, and $g_{t} \approx 0$ implies the swing-up controller is active. This output is fed through a standard switching hysteris function, to avoid rapidly switching on the class boundary, parameters given in the appendix. The swing-up controller can be seen as the policy network from vanilla SAC, the action then is determined by equation (\ref{act}). The parameters for these networks are given in the appendix. The balancing controller is a linear quadratic regulator $C: S \rightarrow A$  about the acrobot's unstable equilibrium. We use the LQR designed by Spong \cite{spong_swing_1994}:

Using 

\[ Q = \begin{pmatrix} 1000 & -500 & 0 & 0 \\ -500 & 1000 & 0 & 0 \\ 0 & 0 & 1000 & -500 \\ 0 & 0 & -500 & 1000\end{pmatrix}, R = \begin{pmatrix} .5 \end{pmatrix}
\]

The resulting control law is: 
\[ u = -Ks \]

with 

\[ K = [-1649.8,  -460.2,  -716.1,  -278.2] \] 

These three functions together form our policy,  $\pi_{\theta}$. Algorithm \ref{alg:rollout} demonstrates how the action is computed at each timestep.

We learn the basin of attraction for the regulator by framing it as a classification problem, our neural network takes as input the current state, and outputs a class prediction between 0-1. A one implying that the LQR is able to stabilize the system, and a zero implying that it cannot. We then define a threshold function $T(s)$, as a criteria for what we consider a successful trial:

\begin{dmath} {T(s) = \norm{s_{t} - gs } < \epsilon_{thr}} \quad \forall t \in \{N_{e}-b, ...,  N_{e} \}  \label{T} 
\end{dmath}

Here $s$ is understood to be an entire trajectory of states, $N_{e}$ is the length of each episode, $e_{thr}$ and $b$ hyper parameters with values given in the appendix. We are following the convention of a programming language here, (\ref{T}) returns one when the inequality holds, and zero otherwise. To gather data, we sample a random initial condition, do a policy roll out using the LQR, and record the value of \ref{T} as the class label.

To train the gating network we minimize the binary cross entropy loss:

\begin{dmath}L^{\text{G}} =  \mathop{\mathbb{E}}_{\gamma} -\left[ c_{w} y_{i}\log(G_{\gamma}(s_{i})) + (1 - y_{i})\log(1 - G_{\gamma}(s_{i})) \right]\end{dmath} 

Where $y_{i}$ is the class label for the ith sample, $c_{w}$ is a class weight for positive examples. we set $c_{w} = \frac{n_{t}}{n_{p}}w $ where $n_{t}$ is the total number of samples, $n_{p}$ is the number of positive examples, and $w$ is a manually chosen weighting parameter to encourage learning a conservative basin of attraction. We found that the learned basin was very sensitive to this parameter, a value of .01 empirically works well. Note that unlike the other losses above, the data here is not computed over a sample but is instead computed over the entire replay buffer. We found the gate was prone to "forgetting" the basin of attraction early in the training otherwise. This also allows us to update the gate infrequently compared to the other networks, and so the total impact on wall clock time is modest.

The second extension is a modification of the replay buffer $D$. We do this by constructing $D$ from two separate buffers, $D_{n}$ and $D_{r}$. Only roll outs that ended in a successful balance (as defined by equation (\ref{T})) are stored in $D_{r}$. The other buffer stores all trials, the same as the unmodified replay buffer. Whenever we draw experience from $D$, with probability $p_{d}$ we sample from $D_{n}$, and with probability $(1-p_{d})$ we sample from $D_{r}$. We found this to speed up learning dramatically, as even with the LQR and a decent gating function in place, the swing-up controller finds the basin of attraction only in a tiny minority of trials.



\begin{algorithm}
\caption{Do-Rollout($G_{\gamma}, \Pi_{\theta}$, K)}
\label{alg:rollout}
\begin{algorithmic}[1]
\State $s = r = a = g = r = \{\}$
\State  Reset environment, collect $s_{0}$
\For    {$t \in \{0, ..., T\} $}
\State  $g_{t} = hyst(G_{\gamma}(s_{t}))$ 
\If {$(g_{t}) == 1$}
\State    $a_{t} = -Ks_{t}$
\Else
\State   Sample $\epsilon_{t}$ from $N(0, 1)$
\State   $a_{t} = \beta\tanh(\mu_{\theta}(s_{t}) + \sigma_{\theta}(s_{t})*\epsilon_{t})$ 
\EndIf
\State   Take one step using $a_{t}$,  collect $\{s_{t+1}, r_{t}\}$
\State   $s = s \bigcup s_{t}$, $r = r \bigcup r_{t}$
\State   $a = a \bigcup a_{t}$,  $g = g \bigcup g_{t}$
\EndFor
\State \bf{return} $s, a, r, g$
\end{algorithmic}
\end{algorithm}

\begin{algorithm}
\label{algo:SSAC}
\caption{Switched Soft Actor Critic}\label{euclid}
\begin{algorithmic}[1]
\State Initialize network weights $\theta ,\phi, \gamma, \rho_{1}, \rho_{2}$ randomly
\State set $\overline \phi = \phi$
\For  {$n \in \{0, ..., N_{e}\} $}
\State $s,r,a,g = \text{Do-Rollout}(G_{\gamma}, \Pi_{\theta}, K)$
\If {$T(s)$}
\State Store $s,r,a$ in $D_{n}$
\EndIf
\State Store $s,r,a$ in $D_{r}$
\State Store $s,g,T(s)$ in $D_{g}$
\If {Time to update policy}
\State sample $s^{r}, a^{r}, r^{r}$ from $D$
\State $\hat Q \approx R + \gamma V_{\overline{\phi}}(S)$
\State $Q^{min} = \min(Q_{\rho_{1}}(s^{r},a^{r}), Q_{\rho_{2}}(s^{r},a^{r}))$
\State $\hat V \approx Q^{min} - \alpha H(\pi_{\theta} (A|S))$
\State Run one step of Adam on $L^{Q}(s^{r}, q^{r}, r^{r})$
\State Run one step of Adam on $L^{\pi}(s^{r})$
\State Run one step of Adam on $L^{V}(s^{r})$
\State   $\overline \phi =  q \overline \phi + (1-q)\phi$
\EndIf
\If {Time to update gate}
\State Run one step of Adam on $L^{G}$ using all samples in $D_{g}$
\EndIf
\EndFor
\end{algorithmic}
\end{algorithm}

\section{Results}

\subsection{Training} 
To train SSAC we first start by training the gate exclusively, using the supervised learning procedure outlined in section \ref{section:SSAC} This allows us to form an estimate of the basin of attraction before we try to learn to reach it. We trained the gate for 1e6 timesteps, and then trained both in parallel using algorithm 2 for another 1e6 timesteps. The policy, value, and Q functions are updated every 10 episodes, and the gate every 1000. The disparity is because, as mentioned earlier, the gate is updated using the entire replay buffer, while all the other losses are updated with one sample batch from the buffer. Hyperparameters were selected by picking the best performing values from a manual search, which are reported in the appendix.

In addition to training on our own version of SAC and Switched SAC we also examined the performance of several algorithms written by OpenAI and cleaned up by the community \cite{stable-baselines}. We examine PPO and TRPO, two popular trust region methods. A2C was included to compare to a non trust region, modern policy gradient algorithm. We also include TD3, which has been shown in the literature to do well on the acrobot and cartpole problems \cite{lillicrap_continuous_2015}

\begin{figure}[h]
\centering
  \includegraphics[scale=.5]{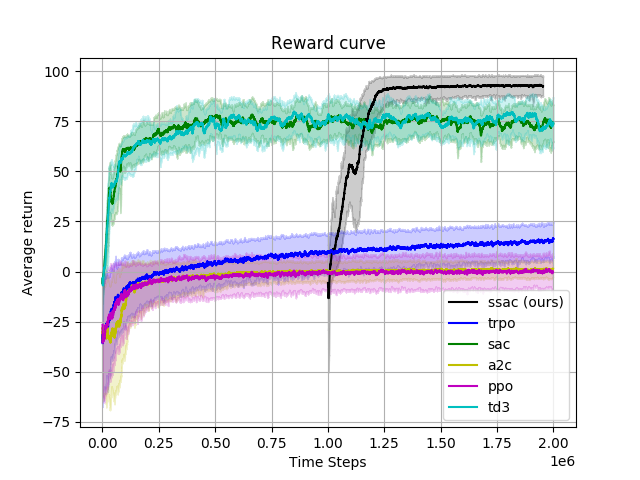}
  \caption{Reward curve for SSAC and the other algorithms we compare to. the solid line is the smoothed average of episode reward, averaged over four random seeds. The shaded area indicates the best and worst rewards at each epoch across the four seeds. SSAC is shown starting later to account for the time training the gating function alone.}
  \label{fig:switched_reward}
\end{figure}

Stable baselines includes hyperparameters that were algorithmically tuned for each environment. For algorithms where parameters for Acrobot-v1  were available we chose those, some algorithms were missing tuned Acrobot-v1 examples, and for those we used parameters for Pendulum-v0, simply because it is another continuous, low dimensional task. Note we don't expect the hyper-parameters to impact the learned policy's score in this case, only how fast learning occurs. Reported rewards are averaged over 4 random seeds. Every algorithm makes 2e6 interactions with the environment. Also note that this project was largely inspired by spending a large amount of time manually tuning these parameters to work on this task (with no success better than what we see here). Figure \ref{fig:switched_reward} shows the reward curve for our algorithm and the algorithms from stable baselines. Table \ref{table:results} shows the mean and standard deviation for the final rewards obtained by all algorithms. 

\begin{center}
\begin{tabu}{| X[l] | X[l] |}
\hline
Algorithm (implementation) & Mean Reward $\pm$ Standard Deviation\\
 \hline
 SSAC (Ours) & \bf{92.12 $\pm$ 2.35}  \\
 \hline
 SAC & 73.01 $\pm$  11.41 \\
 \hline
 PPO  &  0.43 $\pm$ 8.89 \\ 
 \hline
 TD3  & 78.67  $\pm$ 61.85 \\
 \hline
 TRPO  &  17.63 $\pm$ 3.39 \\
 \hline
 A2C  & 2.57 $\pm$ 3.63 \\
 \hline
\end{tabu}
\captionof{table}{Rewards after training for across learning algorithms. This table shows results after 2 million environment interactions}
\label{table:results}
\end{center}

As we can see, for this environment, with the number of steps we have allotted, our approach outperforms the algorithms we compared to, with TD3 making it the closest to our performance. This is a necessarily flawed comparison. These algorithms are meant to be general purpose, so it is unfair to compare them to something designed for a particular problem. But that is part of the point we are making, that adding just a small amount of domain knowledge can improve performance dramatically.

\subsection{Analyzing performance}

\begin{figure}[h!]
\centering
  \includegraphics[scale=.5]{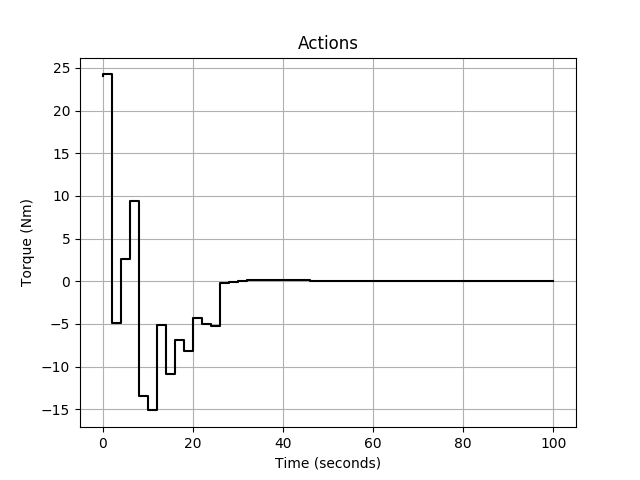}
  \caption{Torque exerted during the sampled episode}
  \label{fig:act}
\end{figure}

\begin{figure}[h!]
\centering
  \includegraphics[scale=.5]{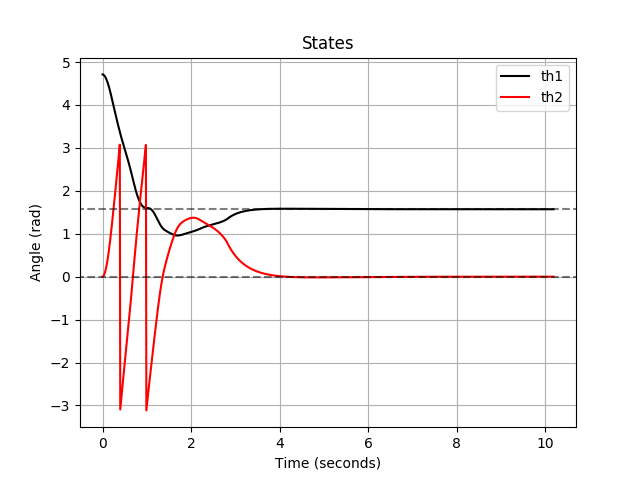}
  \caption{Observations during the sampled episode}
  \label{fig:obs}
\end{figure}

To qualitatively evaluate the performance of our learned agent we examine the behavior during individual episodes. SSAC gives us a deterministic controller (we can set $\epsilon_{t}$ from \ref{act} to zero). We chose the initial condition $s_{0} = (-\pi/2, 0, 0, 0)$ and record a rollout. The actions are displayed in figure \ref{fig:act}, and the positions in \ref{fig:obs}.

\begin{figure}[h]
\centering
  \includegraphics[scale=.35]{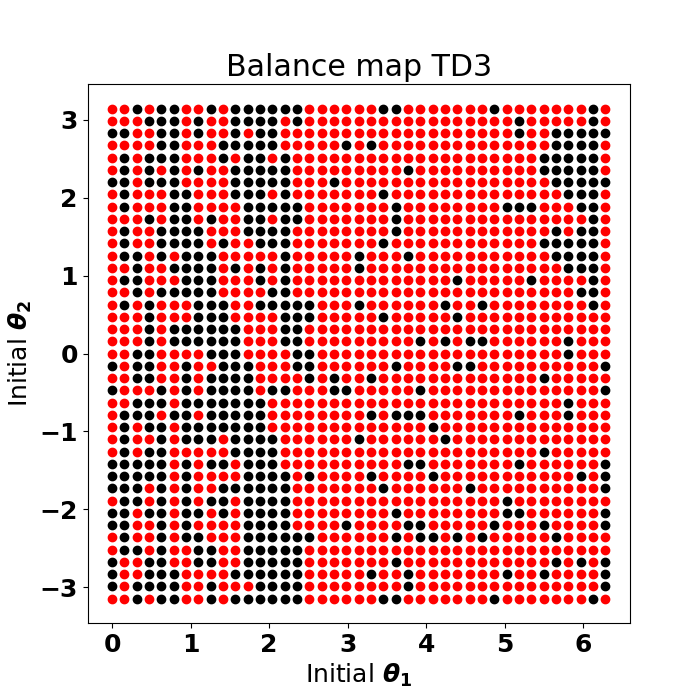}
  \caption{Balance map for TD3, X and Y indicate the initial position for the trial, a black dot indicates that the trial started from that point satisfies equation (\ref{T}), and red indicates the converse. \textbf{when these initial conditions were run for SSAC, it satisfied (11) for every initial condition}}
  \label{fig:td3_map}
\end{figure}

We have also found that despite achieving relatively high rewards, the other algorithms we compare to often fail to meet the balance criteria (11). We often see solutions where the first link is constantly rotating, with the second link constantly vertical. To demonstrate this, as well as to demonstrate our algorithms robustness, we run roll outs with the trained agents across a grid of initial conditions, recording if the trajectory satisfies (11) or not. We compare our method with TD3, which was the best performing model free method we could find on this task. Figure \ref{fig:td3_map} show the results, \textbf{when these initial conditions were run for SSAC, it satisfied (11) for every initial condition}. 

\section{CONCLUSIONS}

We have presented a novel control design methodology that allows engineers to leverage their domain knowledge, while also reaping many of the benefits from recent advances in deep reinforcement learning. In our case study we constructed a policy to swing-up and balance an acrobot while only needing to manually design a linear controller for the balancing task. We believe this method of control will be straightforward to apply to the double or triple cartpole problems, which to our knowledge no model free algorithm is reported as solving. We also think that this general methodology can be extended to more complex problems, such as legged locomotion. In that case the linear controller here could be a nominal walking controller obtained via trajectory optimization, and the learned controller could be a recovery controller to return to the basin of attraction of this nominal controller. 


\bibliographystyle{IEEEtran}
\bibliography{IEEEabrv,IEEEexample}

\section*{APPENDIX}

\subsection*{Hyperparameters}

\begin{center}
\begin{tabu}{ X[2,l] | X[1,l] }
Hyperparameter & Value \\
 \hline
 Episode length ($N_{e}$) & 50  \\ 
 Exploration steps & 5e4 \\
 Initial policy/value learning rate & 1e-3  \\ 
 Steps per update & 500 \\
 Replay batch size & 4096 \\
 Policy/value minibatch size & 128 \\ 
 Initial gate learning rate & 1e-5  \\ 
 Win criteria lookback (b) & 10 \\
 Win criteria threshold ($\epsilon_{thr}$) & .1 \\ 
 Discount ($\gamma$)   & .95 \\
 Policy/value updates per epoch & 4 \\
 Gate update frequency & 5e4 \\
 Needle lookup probability $p_{n}$ & .5 \\ 
 Entropy coefficient ($\alpha$) & .05 \\
 Polyak constant ($c_{py}$)  & .995\\
 Hysteresis on threshold & .9 \\
 Hysteresis off threshold & .5 \\
 \hline
\end{tabu}
\end{center}

\subsection*{Network Architecture}
The policy, value, and Q networks are all made of four fully connected layers, with 32 hidden nodes and Relu activations. The gate network is composed of two hidden layers with 32 nodes each, also with Relu activations, the last output is fed through a sigmoid to keep the result between 0-1.


\addtolength{\textheight}{-12cm}   

\end{document}